\documentclass{article}
\usepackage{spconf,amsmath,graphicx}
\usepackage{hyperref}
\usepackage{caption}
\usepackage{subcaption}
\usepackage{float}
\usepackage{multirow}

\title{YOLOv7 for Mosquito Breeding Grounds Detection and Tracking}
\name{Camila Laranjeira$^1$, Daniel Andrade$^1$, Jefersson A. dos Santos$^{1,2}$ }
\address{$^1$Universidade Federal de Minas Gerais, Brazil\\$^2$University of Sheffield, United Kingdom}
\begin{document}
%
\maketitle
\begin{abstract}
With the looming threat of climate change, neglected tropical diseases such as dengue, zika, and chikungunya have the potential to become an even greater global concern. Remote sensing technologies can aid in controlling the spread of \textit{Aedes Aegypti}, the transmission vector of such diseases, by automating the detection and mapping of mosquito breeding sites, such that local entities can properly intervene. In this work, we leverage YOLOv7, a state-of-the-art and computationally efficient detection approach, to localize and track mosquito foci in videos captured by unmanned aerial vehicles. We experiment on a dataset released to the public as part of the ICIP 2023 grand challenge entitled Automatic Detection of Mosquito Breeding Grounds. We show that YOLOv7 can be directly applied to detect larger foci categories such as pools, tires, and water tanks and that a cheap and straightforward aggregation of frame-by-frame detection can incorporate time consistency into the tracking process.    

\end{abstract}
\section{Introduction}
\label{sec:intro}

Arboviral diseases such as dengue, zika and chikungunya, are among the World Health Organization (WHO) list of Neglected Tropical Diseases (NTD)~\cite{world2023globaltech}, with the mosquito species \textit{Aedes Aegypti} as their main vector. WHO's most recent report cites that climate change might induce an expansion of \textit{Aedes Aegypti}'s geographical range~\cite{world2023globaltech}. Such a claim is under study in the literature, highlighting the adaptability of the mosquito to regions beyond the tropics, provided that temperature keeps rising~\cite{booth2018climate, portner2022climate}. Arboviral diseases are expected to become a global concern in the long term.

Since there is currently no vaccines for most diseases transmitted by \textit{Aedes Aegypti}, the most effective approach is to prevent the mosquito from spreading by finding and eliminating potential breeding grounds, i.e., the accumulation of clean and stagnant water~\cite{lambrechts2012vector}. To that end, the \textit{Automatic Detection of Mosquito Breeding Grounds}~\footnote{\url{https://www02.smt.ufrj.br/~tvdigital/mosquito/challenge/}}, one of the grand challenges at the 2023 International Conference on Image Processing (ICIP), invites researchers to develop object detection and tracking techniques applied to a dataset released by the same authors of the competition~\cite{passos2022automatic}, entitled Mosquito Breeding Grounds (MBG). They encourage researchers to join the fight against arboviral diseases by automating the tiresome work of locating common mosquito foci such as bottles, pools and water tanks, which is often conducted by health agents walking door-to-door~\cite{conas2017norm}.

The aforementioned grand challenge prompts research in the field of remote sensing, since it provides a dataset of annotated videos collected by Unmanned Aerial Vehicles (UAVs), or drones. Labels refer to frame-by-frame detection bounding boxes along with the correspondent object class and instance indices. It presents challenges such as high resolution images as well as large scale variation in regions of interest due to different flight heights and the characteristics of mapped classes (e.g. pools vs bottles). Most prominently, it requires time consistency, since the provided labels are associated with unique object indices persistent throughout the video.

Recent work on remote sensing to detect mosquito breeding grounds relies on state-of-the-art deep learning architectures such as Faster R-CNN~\cite{haddawy2019large, cunha2021water} and YOLOv3~\cite{bravo2021automatic}. ICIP's challenge provides dense labels in time (i.e., frame-by-frame), which can be computationally intensive to automate. Thus in this work we opt to experiment with YOLOv7~\cite{wang2022yolov7}, which has a good trade-off between accuracy and time performance, producing a near real-time approach to detection and tracking of mosquito foci. Thus we supplement the proposition in \cite{bravo2021automatic}, leveraging an improved version of YOLO and adapting cheap and straightforward approaches to aggregate per-frame detections. The latter is an adaptation of \cite{han2016seq}, mainly based on intersection-over-union (IOU) measures and spatio-temporal distances of objects overlapping in time. 

The remaining of this paper is organized as follows. Section \ref{sec:met} describes our proposed methodology, divided into two steps: per-frame object detection and an aggregation of inferences to track unique instances in time. Section \ref{sec:results} is a detailed description of our experiments including all training details, dividing results into raw detections and object tracking. Finally, section \ref{sec:conc} is a brief discussion of our achievements, limitations and future steps.

\section{Methodology}
\label{sec:met}
In this section we outline two separate stages of our proposition. First, we trained YOLOv7 to perform frame-by-frame object detection for all six classes of the MBG database, described in section \ref{sec:dataset}. With the output of that stage for each video, a temporal analysis is performed to assign unique indices for objects that overlap in time, mainly based on IOU and spatio-temporal distances.

\subsection{YOLO Detection}
\label{sec:yolo}
\begin{figure}
    \centering
    \includegraphics[width=0.5\textwidth]{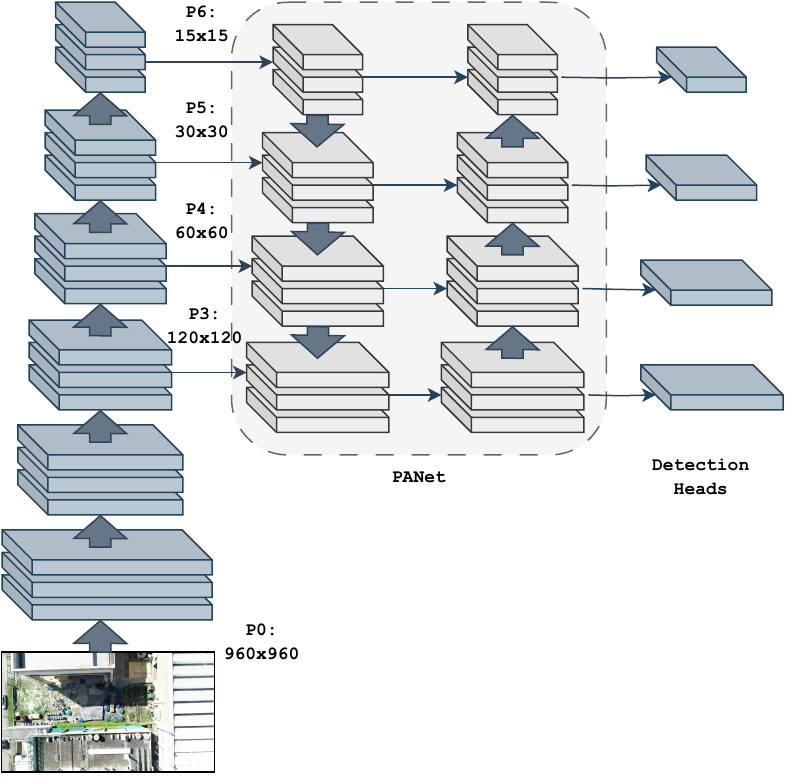}
    \caption{YOLOv7 architecture adapted from ~\cite{maji2022yolo}.}
    \label{fig:yolo}
\end{figure}

The first step is to train an image detection algorithm to work on a frame-by-frame level. YOLOv7 was our choice of architecture since it currently provides one of the best trade-offs between computational cost and detection accuracy~\cite{wang2022yolov7}. The original proposition relies on a fully convolutional backbone to process the image followed by a pyramid matching in 3 different levels of the architecture. We chose, however, to adopt the proposition from \cite{maji2022yolo}, with a 4-level pyramid, as depicted in Figure \ref{fig:yolo}. Since the task at hand demands locating small objects in the scene, our rationale was that adding an earlier level from the architecture would be more adequate. Our greatest concern for this task was to balance computational cost and detection performance, thus a hyperparameter of interest is the input image resolution.

The output provided by the chosen architecture is a spatially structured grid of inferences sized $l \times a \times s \times s$, respectively the number of layers in the pyramid, the number of anchor priors (refer to \cite{redmon2017yolo9000} for details on the anchor-based system) and the output's width and height, which allows to map the output location back to the input image. Each inference is a $(x,y,w,h,conf,c_1,c_2...,c_n)$ vector with $(x,y,w,h)$ bounding box coordinates, a confidence score, and a probability $c$ for each of the $n$ classes.

Another important aspect of YOLO is the non-maximum suppression stage, where low confidence inferences are dropped given a confidence threshold, and spatially overlapping bounding boxes are removed according to an additional threshold for IOU, favoring higher confidence boxes. 

\subsection{Time consistency}
\label{sec:track}
Given frame-by-frame inferences over a video, the next challenge is associating bounding boxes corresponding to the same instance over time. Our approach relies on a simple assumption that two frames not distant in time would produce high IOU for boxes referring to the same instance, as proposed in \cite{han2016seq}. To account for false negatives, in which an object might be missed for a few frames, we define a time threshold $t$ for instance candidates. 

In other words, in the first frame, all objects spotted are assigned an unique index. From this point forward, we maintain a record of all instances spotted in the last $t$ frames, serving as candidates for detections at $t+1$. Thus, given a threshold for IOU and the objects's classes, every object at $t+1$ may be matched with an existing instance of equal class and maximizing IOU. Objects not matched with any instance are assigned novel indices.

\section{Experiments and Results}
\label{sec:results}

\subsection{Mosquito Breeding Grounds (MBG) Database}
\label{sec:dataset}
The dataset consists of 13 high resolution videos (3840 to 4096 horizontal pixels) with different durations (23'' up to 5'27'') and camera heights (10m to 40m), and fixed frame rate of 24 fps. It is composed with a total of 6 classes, namely bucket, water tanks, bottle, pool, tire and puddle. The number of unique instances also varies widely, with over 700 water tanks and only 3 puddles. The authors provided two separate sets of data, with videos 01, 02, 05, 09, and 13 exclusively assigned to the test set. All training in this work is done with the remaining videos, and we report results on the test split. 

\begin{figure}[ht]
    \centering
    \includegraphics[width=0.41\textwidth]{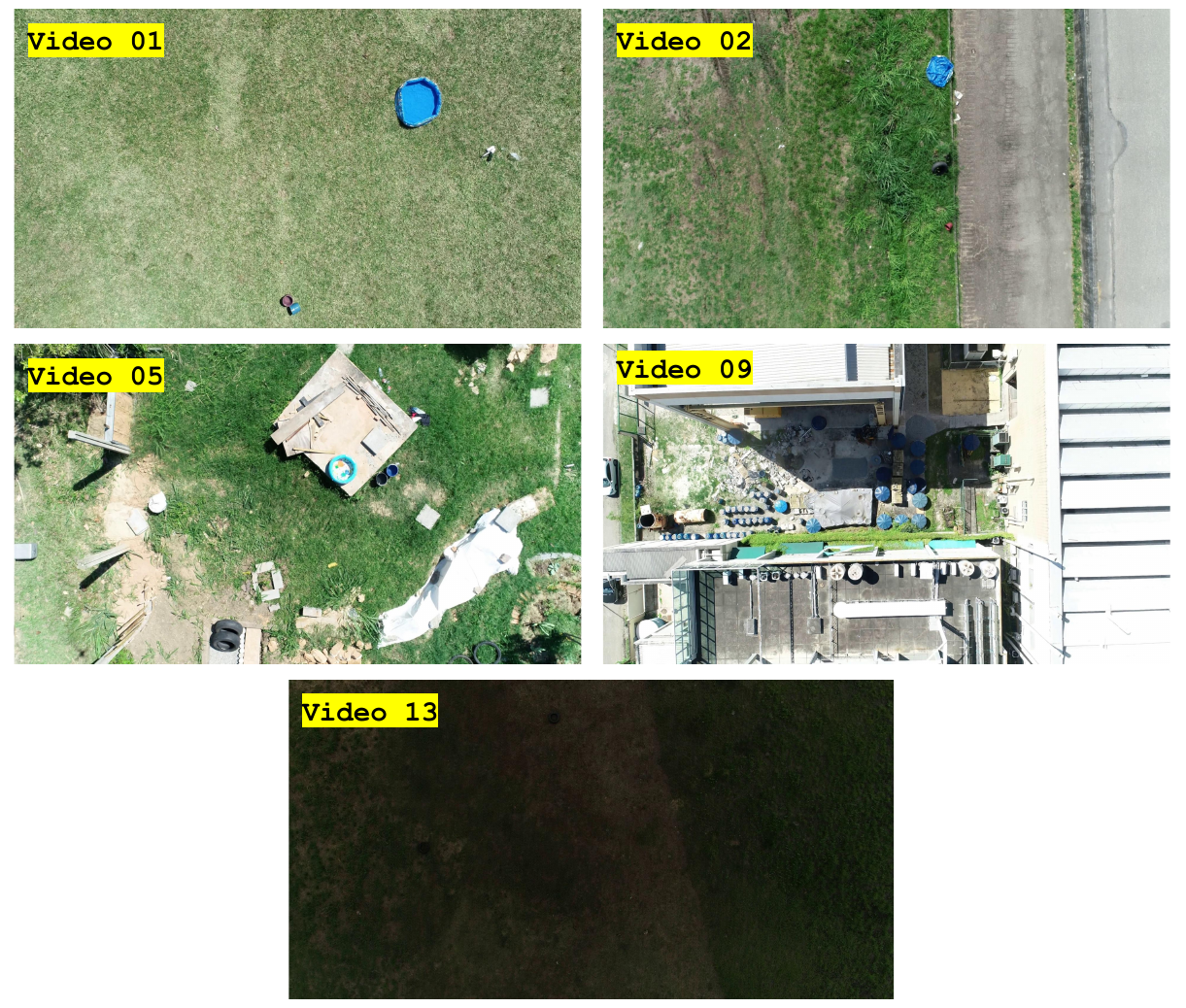}
    \caption{A single frame from each video of the test split.}
    \label{fig:frames}
\end{figure}

To better understand the large variability of the dataset, we selected a single frame from each instance as depicted in Figure \ref{fig:frames}. Notice that in some instances the scene can be crowded or sparse with objects of interest, as well as the large variability of light conditions.

\subsection{Hyperparameters and Training}
After empirical experiments, YOLOv7 was set with input resolution of $960 \times 960$. Values lower than that would compromise even the detection of substantially large objects, like tires and water tanks. From a model pretrained on COCO~\cite{lin2014microsoft}, we freezed YOLO's backbone training only layers related to pyramid matching and detection heads. The architecture was fine-tuned for 50 epochs and batch size of $16$ with the ADAM optimizer. Other hyperparameters were kept as originally set in the code released by \cite{maji2022yolo}~\footnote{\url{https://github.com/TexasInstruments/edgeai-yolov5/tree/yolo-pose}}, with initial learning rate of $10^{-4}$ and weight decay of $5\times10^{-5}$.

For YOLO's non-maximum suppression, both IOU and the confidence threshold were set to $0.5$. And finally, for time consistency, the time threshold was set to $t=45$ frames and $IOU=0.1$. We chose a low IOU and large time tolerance after noticing there was a significant amount of instances where our model would miss smaller objects like buckets and tires for over a second of footage.

\subsection{Results}

\begin{figure}[H]
     \centering
     \begin{subfigure}[b]{0.22\textwidth}
         \centering
         \includegraphics[width=\textwidth]{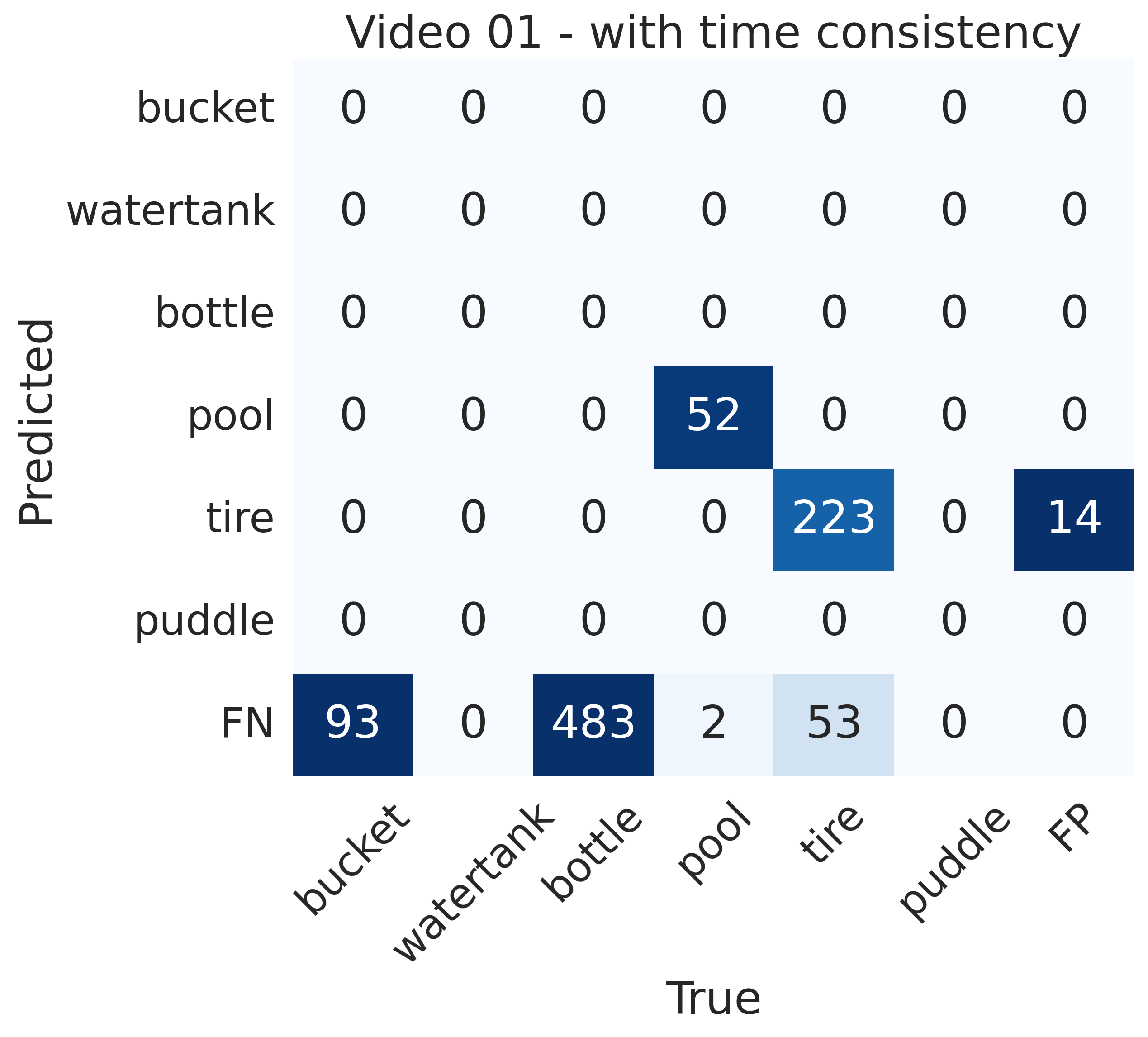}
     \end{subfigure}
     \hfill
     \begin{subfigure}[b]{0.25\textwidth}
         \centering
         \includegraphics[width=\textwidth]{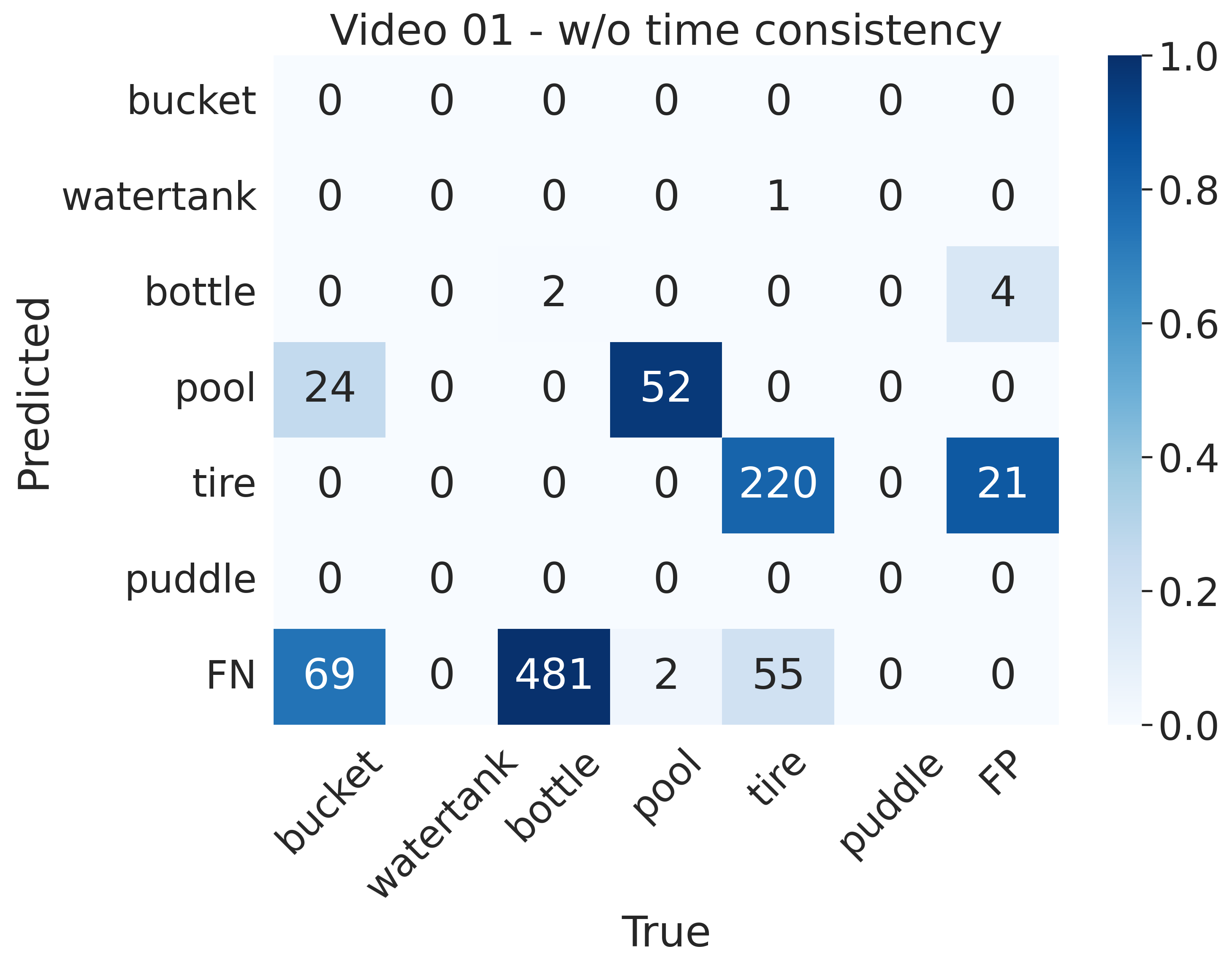}
     \end{subfigure}
     \begin{subfigure}[b]{0.22\textwidth}
         \centering
         \includegraphics[width=\textwidth]{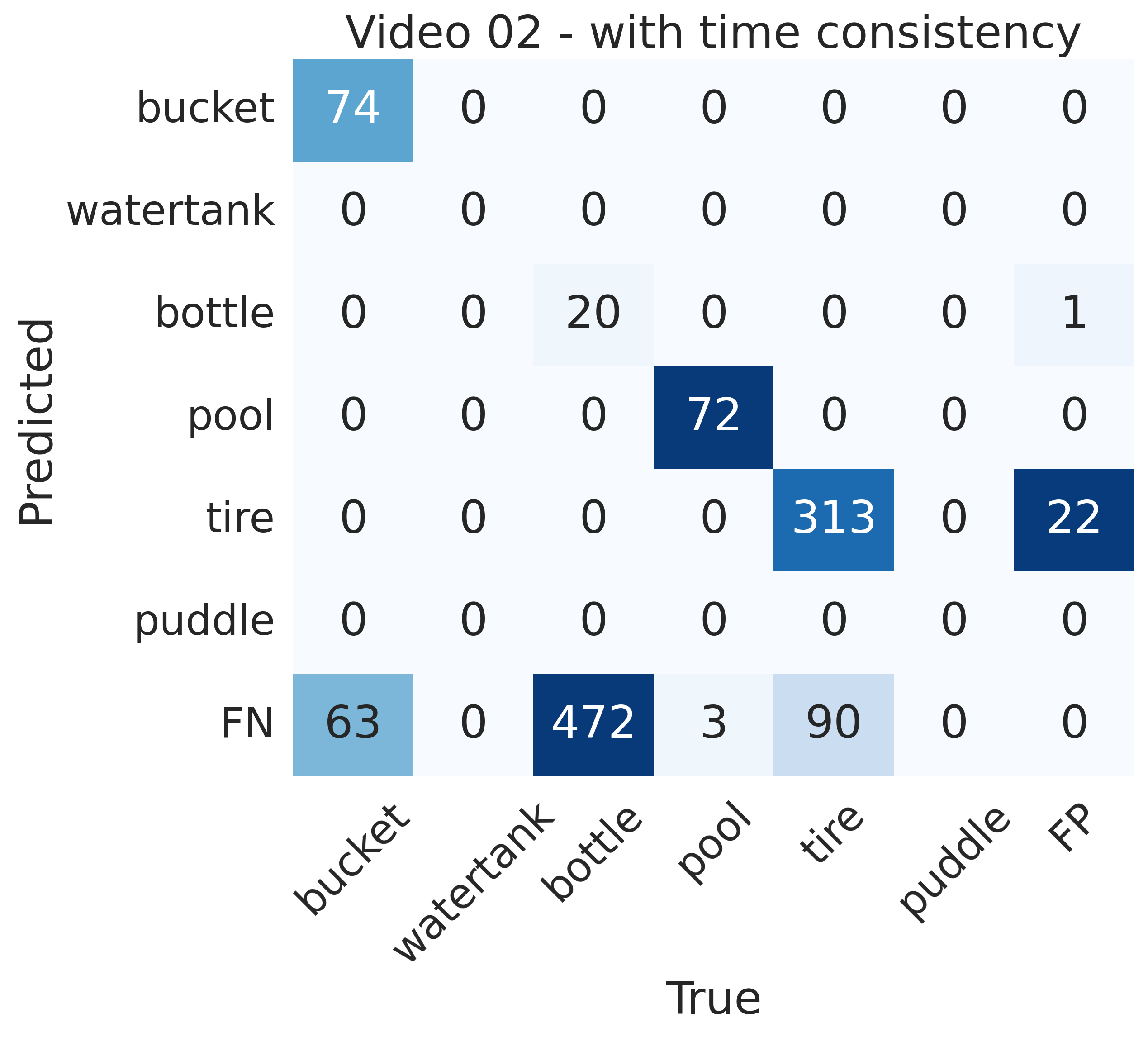}
     \end{subfigure}
     \hfill
     \begin{subfigure}[b]{0.25\textwidth}
         \centering
         \includegraphics[width=\textwidth]{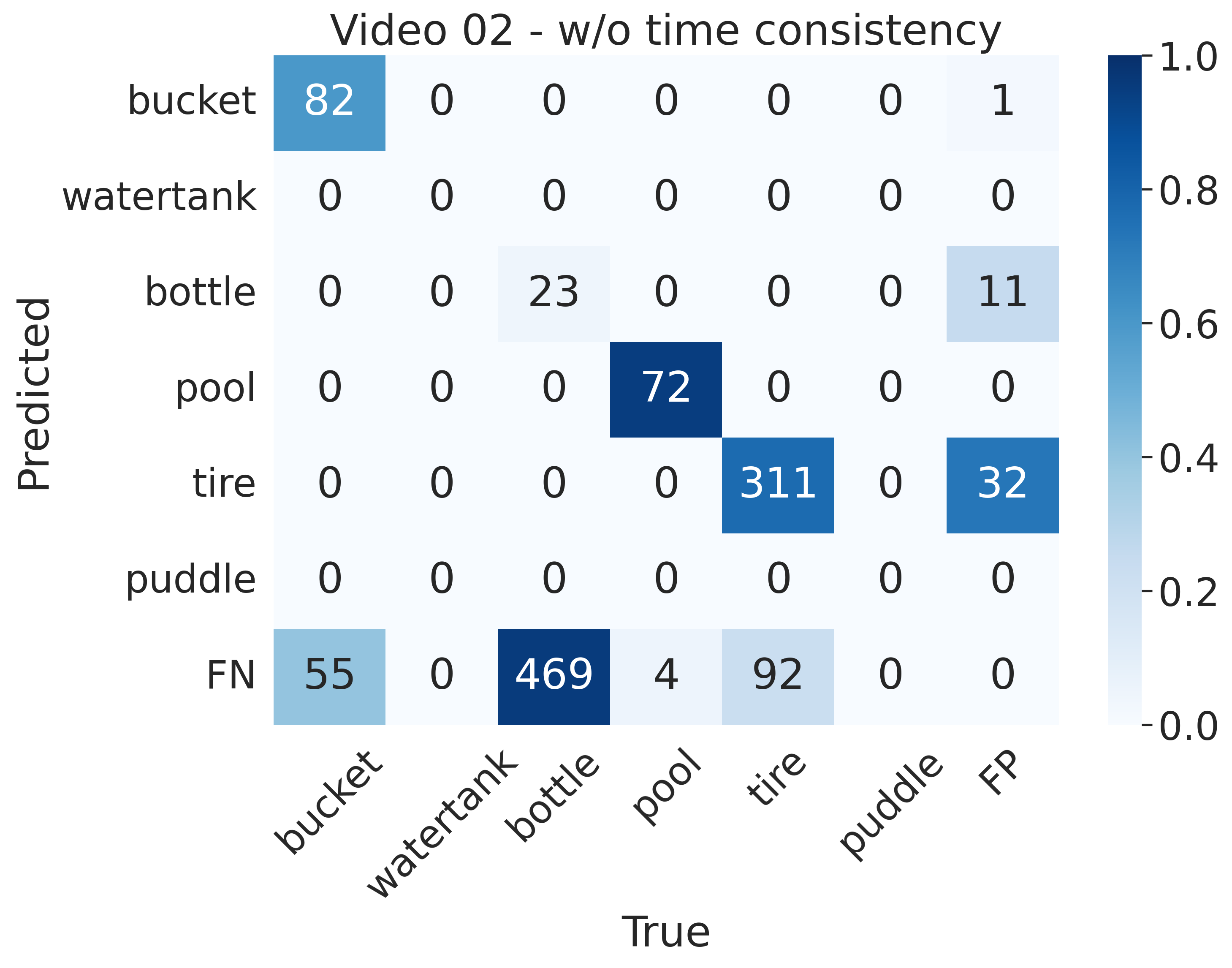}
     \end{subfigure}
     \begin{subfigure}[b]{0.22\textwidth}
         \centering
         \includegraphics[width=\textwidth]{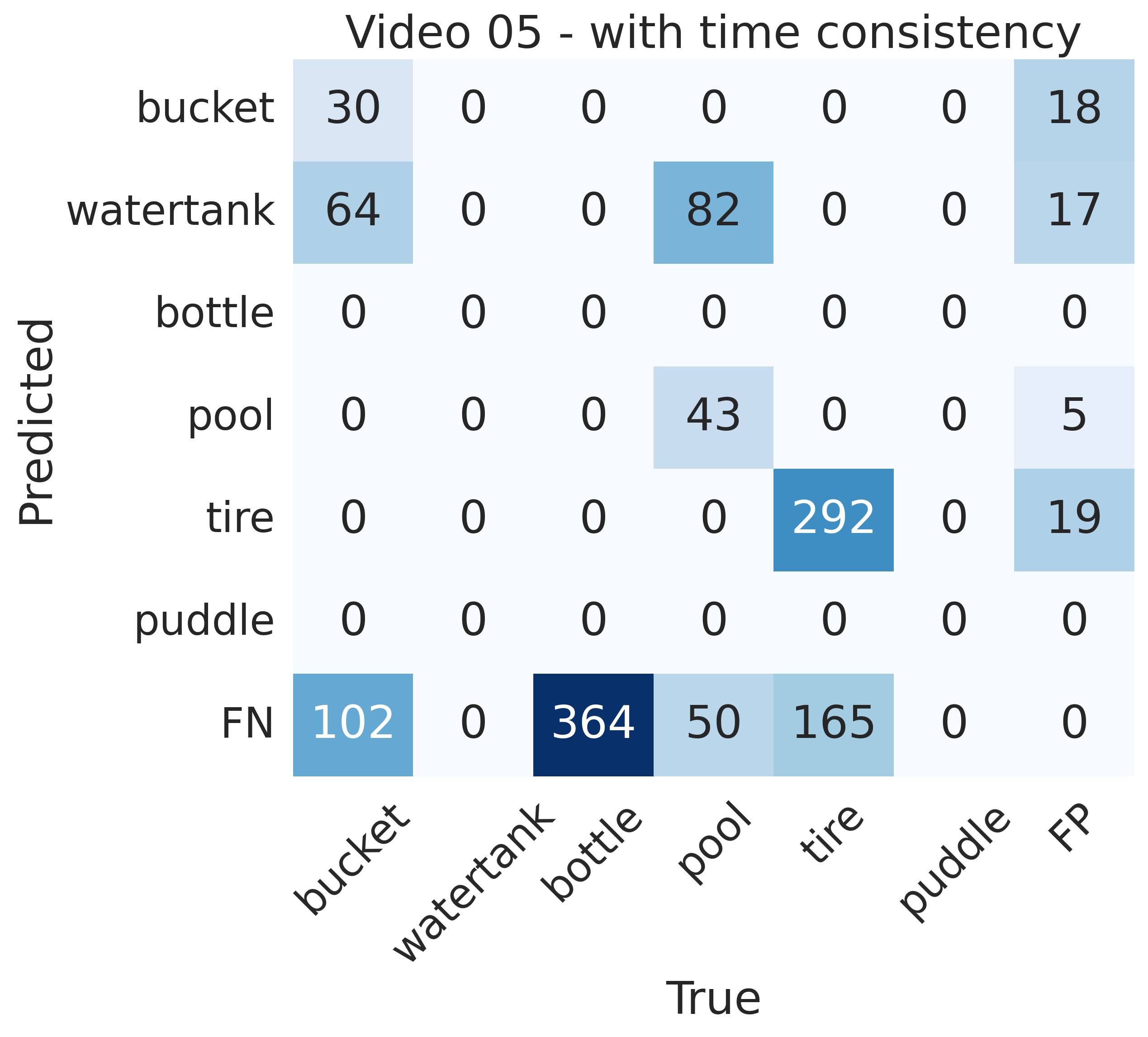}
     \end{subfigure}
     \hfill
     \begin{subfigure}[b]{0.25\textwidth}
         \centering
         \includegraphics[width=\textwidth]{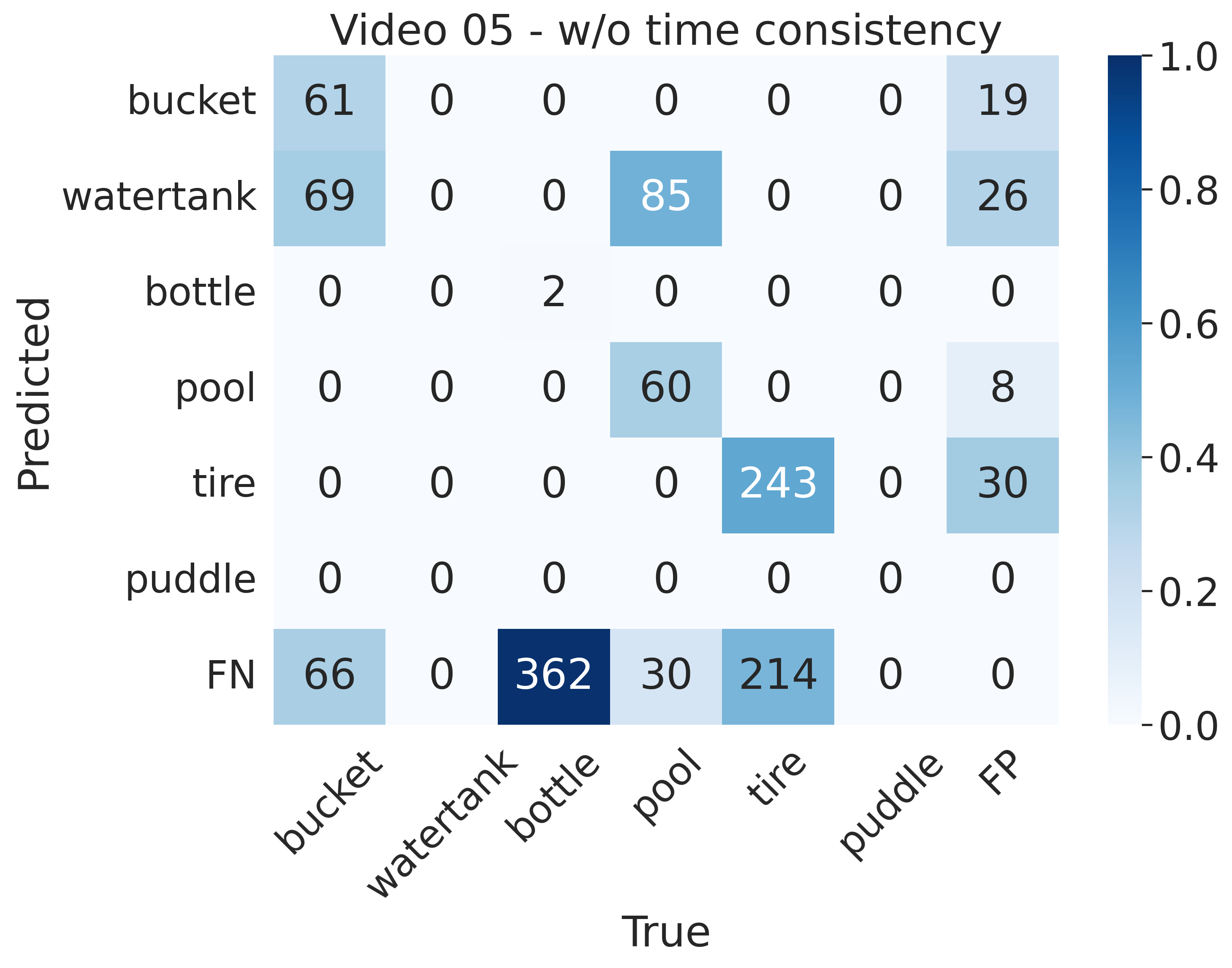}
     \end{subfigure}
     \begin{subfigure}[b]{0.22\textwidth}
         \centering
         \includegraphics[width=\textwidth]{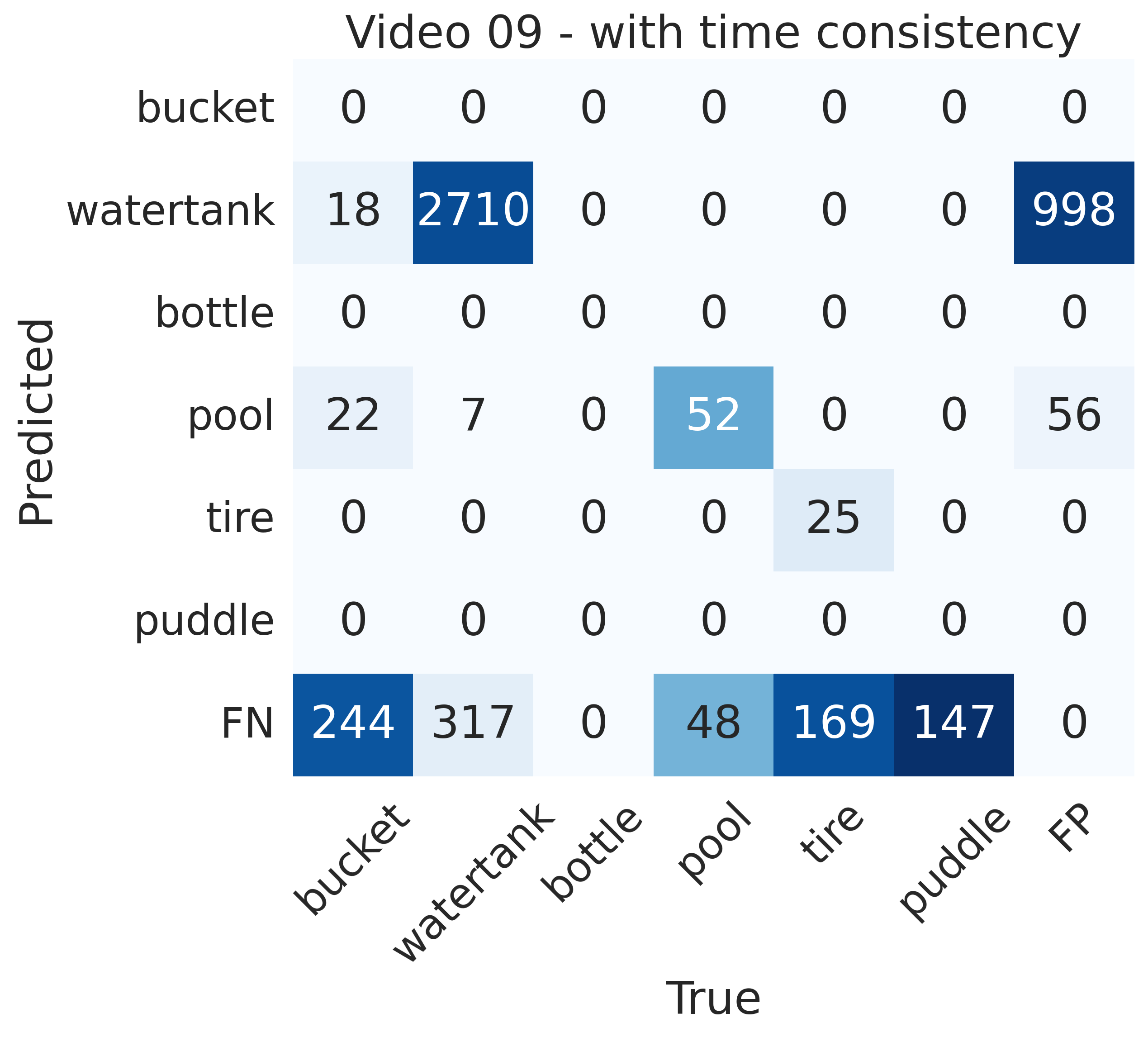}
     \end{subfigure}
     \hfill
     \begin{subfigure}[b]{0.25\textwidth}
         \centering
         \includegraphics[width=\textwidth]{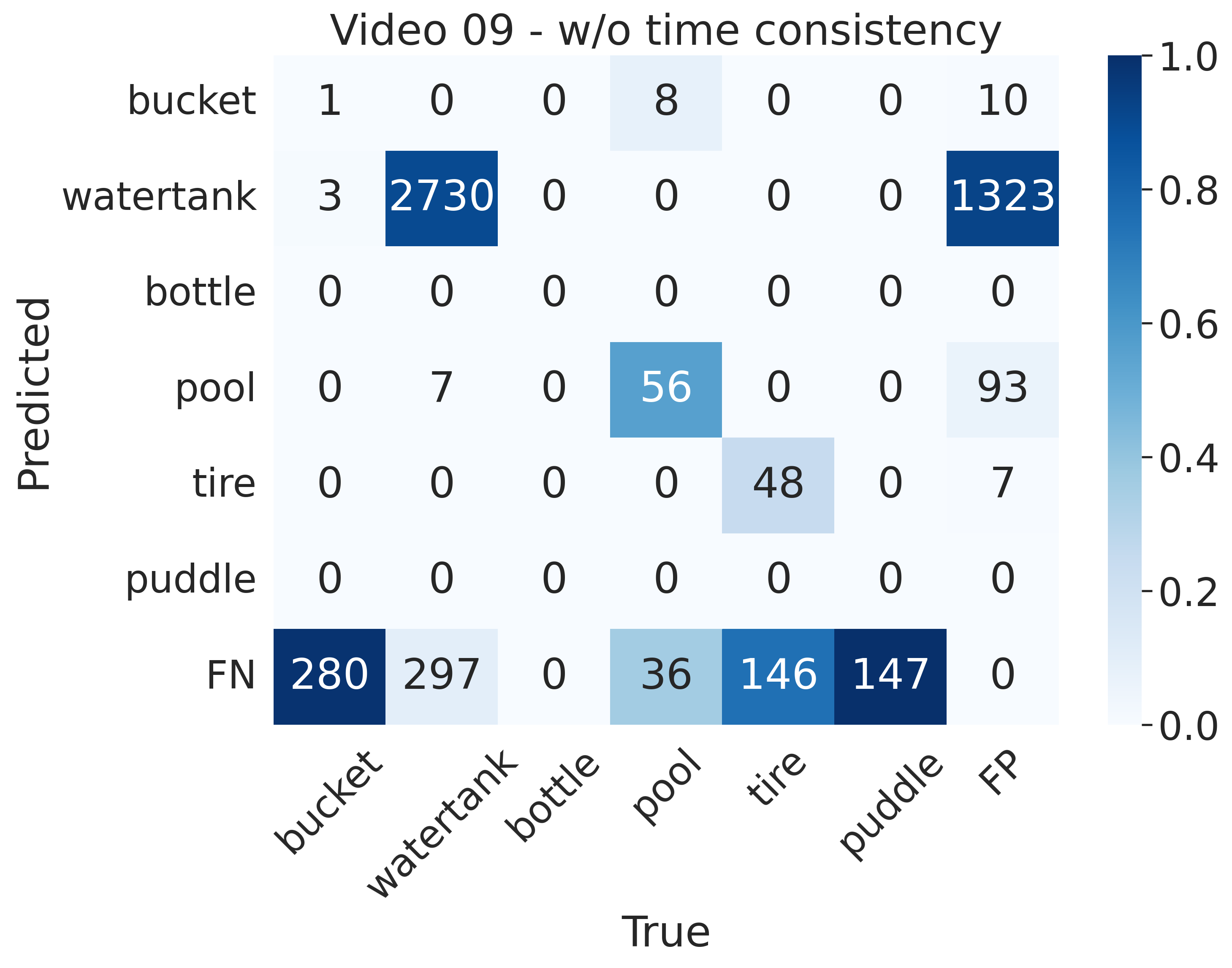}
     \end{subfigure}
     \begin{subfigure}[b]{0.22\textwidth}
         \centering
         \includegraphics[width=\textwidth]{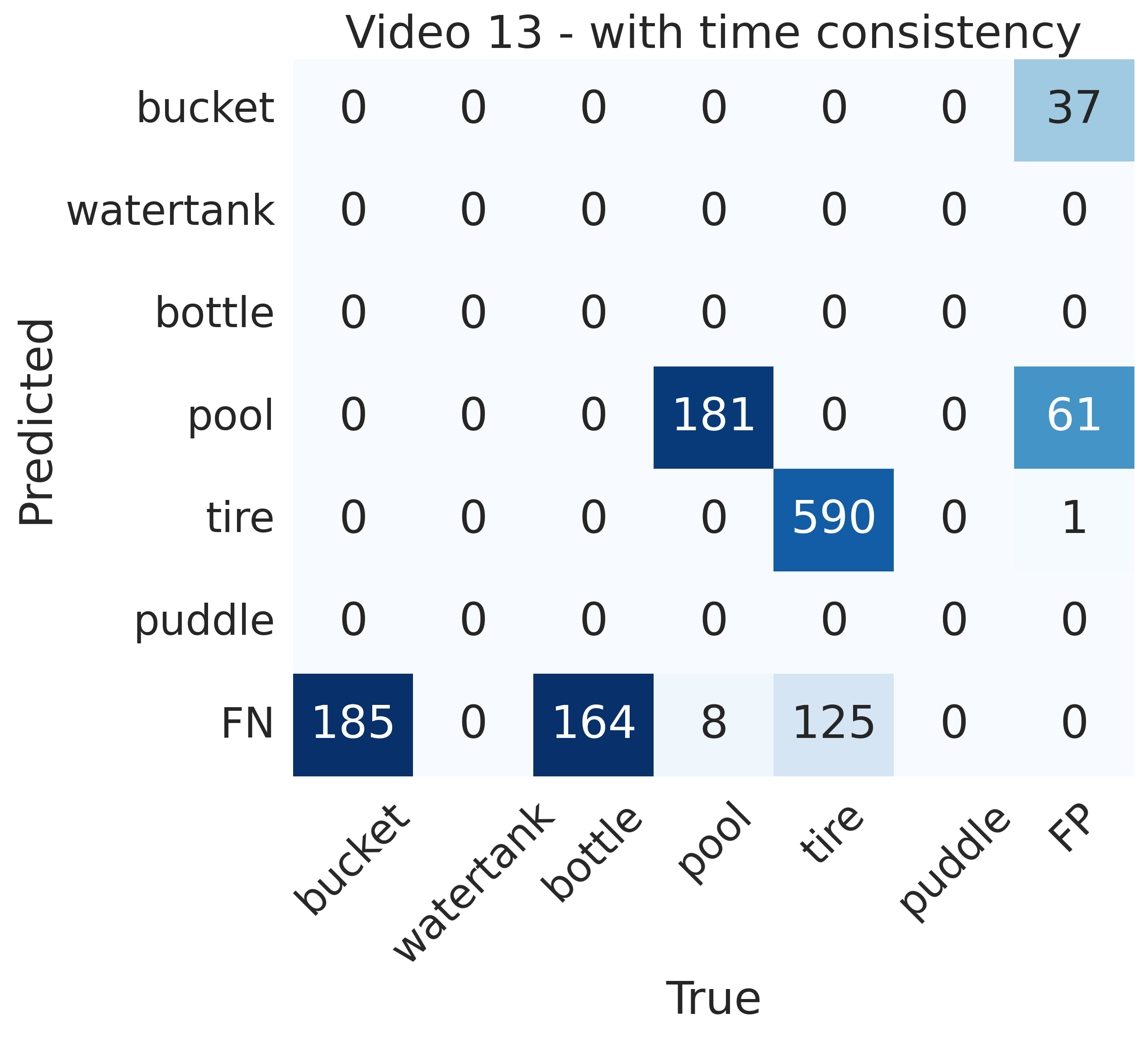}
     \end{subfigure}
     \hfill
     \begin{subfigure}[b]{0.25\textwidth}
         \centering
         \includegraphics[width=\textwidth]{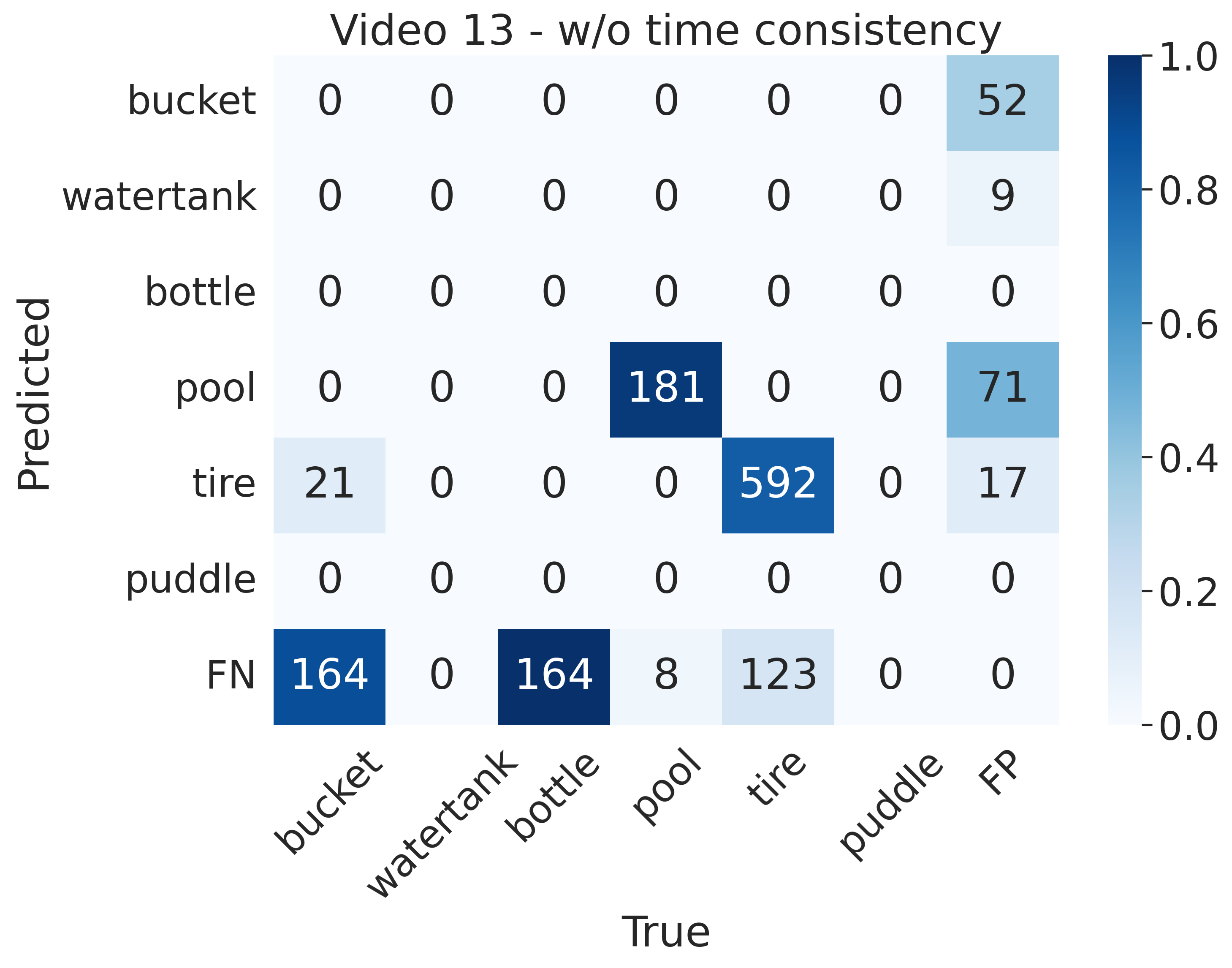}
     \end{subfigure}
        \caption{Frame-by-frame confusion matrices for each video. On the left raw detection values are depicted, and the right column presents results after time consistency is added. While numbers are raw classification counts, colors represent y-normalized values according to the colorbar.}
        \label{fig:results}
\end{figure}

Figure \ref{fig:results} compiles the results from our entry to ICIP's grand challenge with a confusion matrix for each test video, with and without time consistency. The most prominent behaviour is that larger objects like pool, water tank and tire have the greatest accuracies in all samples, followed by the bucket class, with worse behaviour. Finally, bottles and puddles are hardly ever detected. For the bottle class we believe scale is the issue, since this is the smallest of objects from the dataset, although further experiments are required to confirm that. As for puddles, we did not intervene to balance the traning data, which might cause the worse behaviour.

Regarding time consistency, it is noticeable that it decreases the number of false positive detections, which happen sparsely throughout the video, with little compromise to true positive detections. For videos $01$ and $02$, which are the less challenging of the test split, we can see that larger objects like pools and tires achieve over $95\%$ and $80\%$ accuracy respectively. Further analysis is required for the bucket class in these two videos to explain the vast difference in behaviour.

\newpage

Videos $05$ and $09$ are more crowded, posing a greater challenge for YOLO, specially for the tire class. We believe this is due to YOLO's reliance on dividing the image into a grid of size $s \times s$, assigning the ground truth of each object to a few cells in the grid. Thus overlapping objects are often missed, as it is clear in the results regardless of time consistency. Further analysis is required to support this conjecture. Video $09$ is also the only one with water tanks in the test set, and although the model achieves over $90\%$ recall, precision is compromised by the large number of false positives. Lighting conditions did not seem a big issue for our approach. Although video $13$ is particularly dark, the model achieved over $80\%$ and $90\%$ accuracy for classes tire and pool respectively.


To assess the quality of our proposition from a spatio-temporal perspective, we need to associate labelled instances to predicted ones in order to derive classification metrics. For a single video, we take as reference the first frame an instance $L$ was labelled as $fl0$, with the respective bounding box $l_{fl0}$. The same goes for a predicted instance $P$, first sighted in frame $fp0$ with a bounding box $p_{fp0}$. Their last sight are associated with frames $fln$ and $fpn$. The following criteria is used to match instances and predictions:
\begin{enumerate}
    \item instances $L$ and $P$ refer to the same object class;
    \item $|fl0 - fp0| < 45$ frames, and $|fln - fpn| < 45$ frames;
    \item $IOU(l_{fl0}, p_{fp0}) \geq 0.1$ and $IOU(l_{fln}, p_{fpn}) \geq 0.1 $;
    \item cosine distance between the displacement vectors of both instances is $dcos_{lp} \leq 1e^{-2}$.
\end{enumerate}
Note that the thresholds were empirically set, and manually verified, for the sake of providing a spatio-temporal metric. For a detailed understanding of item 4, given a labelled instance $L$ consider the coordinates from its first and last sight  as $(xl_0, yl_0)$ and $(xl_n, yl_n)$. The displacement vector $\vec{vl}$ is

\vspace{-5mm}
\begin{align*}
    \vec{vl}_x = xl_n - xl_0 \\
    \vec{vl}_y = yl_n - yl_0.
\end{align*}

The same goes for a predicted $p$ and its displacement vector $\vec{vp}$. Note that $(x,y)$ coordinates refer to the top left of a bounding box. Considering $cos(\vec{v0}, \vec{v1})$ as a function for the cosine distance between vectors $v0$ and $v1$, $dcos_{lp}$ is 

\vspace{-5mm}
\begin{align*}
    dcos_{lp} &= cos(\vec{vl},\vec{vp}). 
\end{align*}


Table \ref{tab:track} presents the resulting metrics for each video. Predictions matched with labelled instances are considered true positives. False positives indicate a prediction that did not match any labelled instance by any of the 4 established criteria. Finally, false negatives refer to labelled instances not matched to any prediction. These metrics aid us in understanding important aspects, such as the high number of true positive instances for the class tire, even for crowded scenes (videos $05$ and $09$). It is also noteworthy that our model had complete recall for the class water tank, finding all 39 instances in video $09$, although precision is an issue.

Another limitation is that the pool class, which seemed to perform well on a frame-by-frame analysis, in terms of unique instances presents a high number of false positives and false negatives in crowded videos. The table also reinforces our limited capacity to find instances of bottles and puddles, as well as the unstable behaviour for the bucket class.


\begin{table}[h]
    \centering
    \begin{tabular}{llccccc}
         \multicolumn{2}{c}{\textbf{Videos}} & \textbf{01} & \textbf{02} & \textbf{05} & \textbf{09} & \textbf{13} \\ \hline
        \multirow{ 3}{*}{Bucket} & TP & 0 & 2 & 2 & 0 & 1 \\
                                 & FP & 0 & 0 & 0 & 0 & 2 \\
                                 & FN & 2 & 0 & 2 & 3 & 1 \\ \hline
        \multirow{ 3}{*}{Watertank} & TP & 0 & 0 & 0 & 39 & 0 \\
                                 & FP & 0 & 0 & 7 & 59 & 2 \\
                                 & FN & 0 & 0 & 0 & 0 & 0 \\ \hline
        \multirow{ 3}{*}{Bottle} & TP & 0 & 2 & 0 & 0 & 0 \\
                                 & FP & 0 & 0 & 0 & 0 & 0 \\
                                 & FN & 9 & 5 & 8 & 0 & 2 \\ \hline
        \multirow{ 3}{*}{Pool}   & TP & 1 & 1 & 1 & 1 & 2 \\
                                 & FP & 0 & 0 & 1 & 5 & 3 \\
                                 & FN & 0 & 0 & 3 & 0 & 0 \\ \hline
        \multirow{ 3}{*}{Tire}   & TP & 5 & 6 & 6 & 1 & 7 \\
                                 & FP & 1 & 1 & 0 & 0 & 2 \\
                                 & FN & 1 & 0 & 4 & 1 & 0 \\ \hline
        \multirow{ 3}{*}{Puddle} & TP & 0 & 0 & 0 & 0 & 0 \\
                                 & FP & 0 & 0 & 0 & 0 & 0 \\
                                 & FN & 0 & 0 & 0 & 2 & 0 \\ \hline
    \end{tabular}
    \caption{Tracking unique objects in each video. TP: true positive, FP: false positive, FN: false negative.}
    \label{tab:track}
\end{table}

\vspace{-5mm}

\section{Conclusion}
\label{sec:conc}
We developed a straightforward approach for detection and tracking of mosquito breeding grounds, based on a state-of-the-art object detection architecture, namely YOLOv7. With a simple yet effective approach to add time consistency, based on the IOU of objects overlapping in time, we managed to have satisfactory results in the test set for classes such as water tanks, pools and tires. Those are in fact the most explored classes when it comes to locating mosquito foci~\cite{cunha2021water, passos2022automatic}. 

This is a step towards robust models to monitor critical areas most affected by arboviral diseases. We expect to deploy an improved version of our proposition in an ongoing partnership in Brazil between the city of Campinas and the Federal University of Minas Gerais~\footnote{\url{http://patreo.dcc.ufmg.br/2021/09/01/wildpixels/}}. Improvement efforts should be dedicated to detecting smaller and/or overlapping objects, specially in crowded scenes, which were the most challenging scnearios for our proposition.


\section*{Acknowledgment}

This research was partially financed by the Coordenação de Aperfeiçoamento de Pessoal de Nível Superior (CAPES), Fundação de Amparo à Pesquisa do Estado de Minas Gerais (FAPEMIG), CNPq (306955/2021-0),  and the Serrapilheira Institute (grant R-2011-37776).



\bibliographystyle{IEEEbib}

\end{document}